%% file: neurips_2026.tex
\documentclass{article}

\usepackage[preprint]{neurips_2026}
\usepackage[utf8]{inputenc} 
\usepackage[T1]{fontenc}    
\usepackage{microtype}      

\usepackage{mathtools}      
\usepackage{amssymb}        
\usepackage{amsthm}
\usepackage{nicefrac}       

\usepackage[table]{xcolor}  

\usepackage{graphicx}
\usepackage{svg}

\usepackage{caption}
\usepackage{subcaption}

\usepackage{booktabs}       
\usepackage{makecell}
\usepackage{diagbox}

\usepackage{float}
\usepackage{algorithm,algorithmic}
\usepackage{multicol}
\usepackage{enumitem}

\usepackage{pifont}
\usepackage{comment}

\usepackage{hyperref}

\title{\textit{Z-Cache}: Accelerating Diffusion Transformers via  Self-Reflection}

\author{%
  \textbf{Zegang Cheng\textsuperscript{1,*} \quad Zhikai Wang\textsuperscript{2,*} \quad Jiacheng Liu\textsuperscript{1} \quad Xiaobing Tu\textsuperscript{3}} \\
  \textbf{Chang Zhou\textsuperscript{1,\textsection} \quad Junjie Chen\textsuperscript{1,\textdagger} \quad Yue Ma\textsuperscript{1} \quad Zhiyuan Ma\textsuperscript{4}} \\
  \textbf{Jinkui Ren\textsuperscript{3} \quad Likai Zou\textsuperscript{3} \quad Linfeng Zhang\textsuperscript{1,\textdaggerdbl}} \\[0.4em]
  {\normalfont\normalsize\textsuperscript{1}Shanghai Jiao Tong University\quad\textsuperscript{2}Southern University of Science and Technology} \\
  {\normalfont\normalsize\textsuperscript{3}Alibaba Cloud\quad\textsuperscript{4}Huazhong University of Science and Technology} \\[0.3em]
  {\normalfont\normalsize\textsuperscript{*}Equal contribution.\quad\textsuperscript{\textdagger}Work done during an internship.\quad\textsuperscript{\textdaggerdbl}Corresponding author.} \\
  {\normalfont\normalsize\textsuperscript{\textsection}The correct name is Chang Zou; Chang Zhou was entered in error.}
}

\begin{document}

\maketitle

\begin{figure*}[h]
    \vspace{-0.3cm}
    \centering
    \includegraphics[width=1\textwidth]{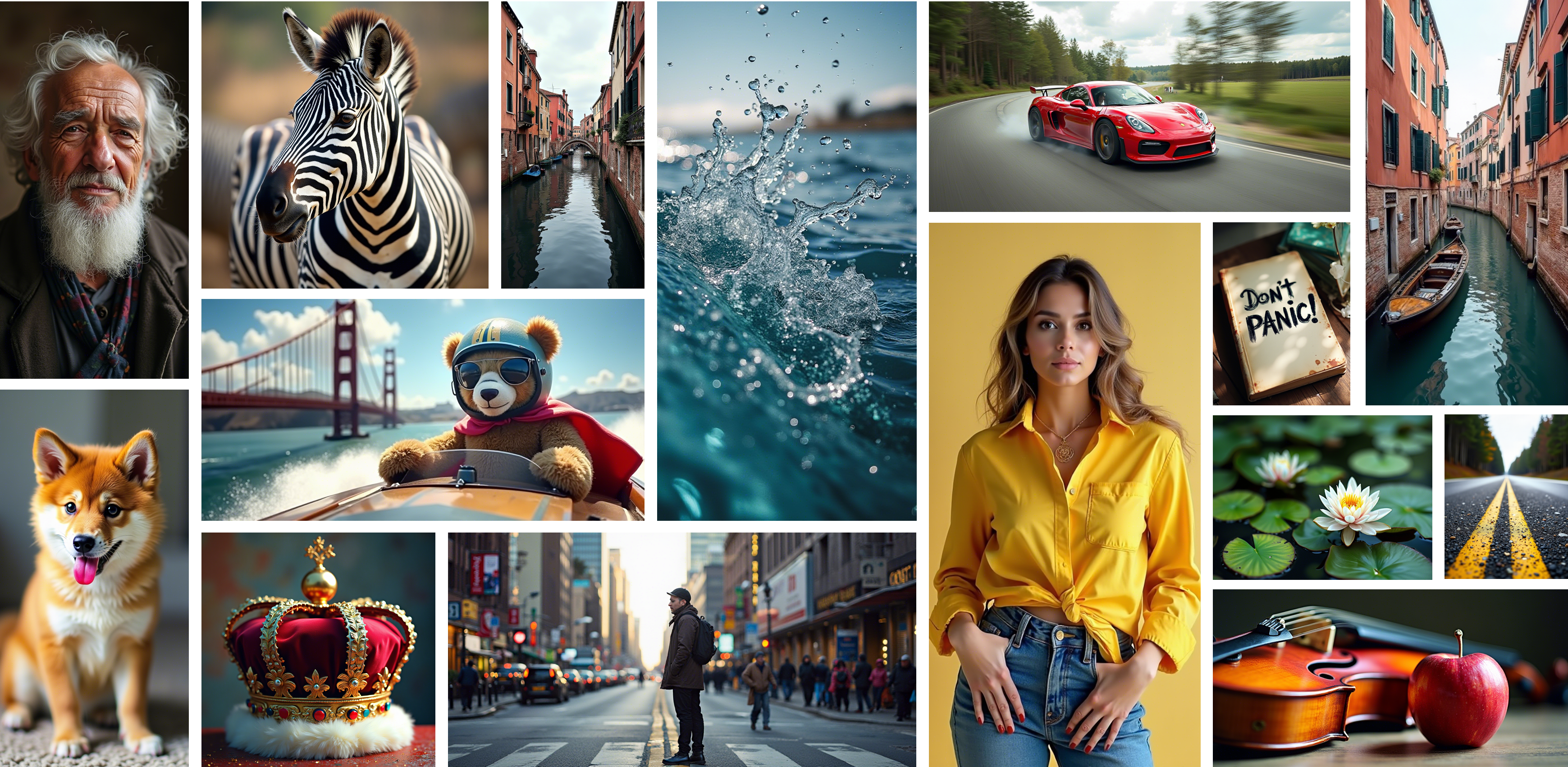}
    \caption{Images sampled by 6.22$\times$ accelerated FLUX.1-dev with \textit{Z-Cache}. }
    \label{fig:showcase}

\end{figure*}
\begin{abstract}
Diffusion transformers have become the most powerful models for visual generation, but still suffer from massive computation costs. To solve this problem, feature caching has been proposed to cache the features of diffusion models in the previous computation steps and then reuse them in the following caching steps, which brings significant acceleration but also degradation in generation quality. To address this problem, this paper proposes Z-cache as a feature caching method that can maintain high-quality generation through self-reflection. Concretely, we observe that \emph{the error from feature caching tends to be sharply reduced after each full computation.} Based on this observation, \textit{Z-Cache} is designed to first predict the features in the future caching steps and then perform a full computation. After that, \textit{Z-Cache} returns to the caching steps and re-predicts them based on the previous and the current computation steps, which brings correction in features.  Experiments demonstrate that with \textit{Z-Cache}, diffusion transformers achieve comparable generation quality to the original model  but with faster inference speed, for instance, \textcolor{black}{\textbf{6.22}}$\times$ acceleration on FLUX-dev for text-to-image generation. \emph{Our code is released on GitHub}.

\begin{figure}[h]
\vspace{-0.2cm}
    \centering
    \includegraphics[width=1\linewidth]{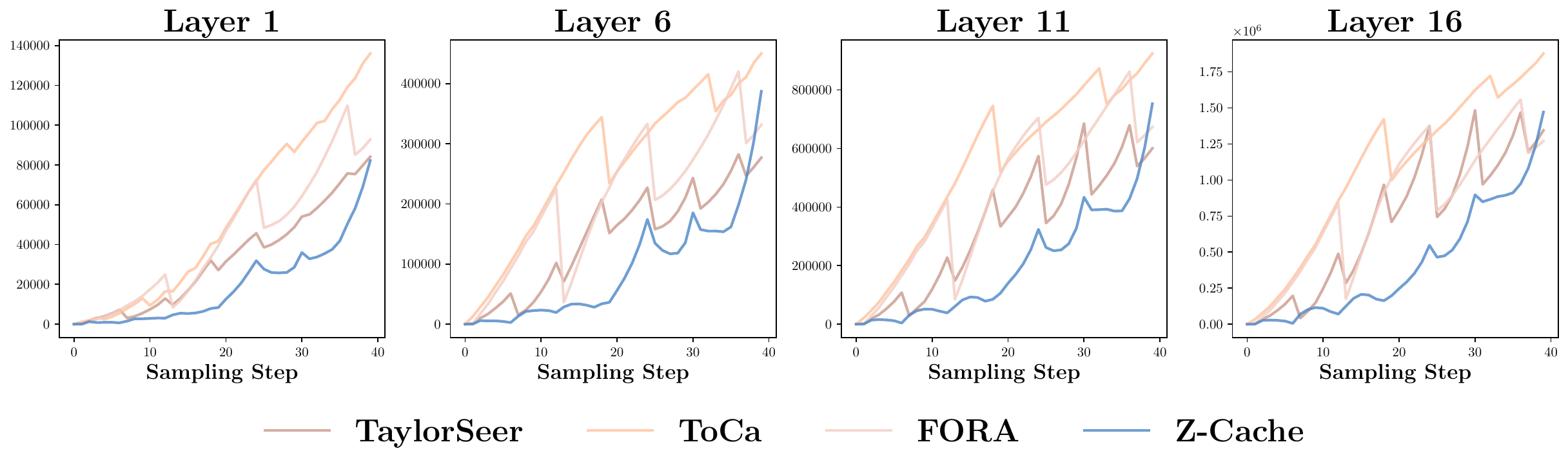}
    \caption{\textbf{Visualization of the per-layer L1 error from feature caching.} It is observed that there exists a periodic decline in L1 error, which is caused by the full-computation step in feature caching. The proposed \textit{Z-Cache} utilizes this phenomenon and derives fewer cumulative errors.
}
    \label{fig:layer-error}
    \vspace{-0.3cm}
\end{figure}

\end{abstract}

\input{sec/intro}
\input{sec/related_work}
\input{sec/method}
\input{sec/experiments}

\input{sec/conclusion}

\renewcommand{\acksection}{\section*{Acknowledgment}}
\begin{ack}
This paper was partially sponsored by WUYING - Alibaba Cloud.
\end{ack}

\clearpage

\bibliography{references}
\bibliographystyle{plainnat}

\clearpage
\appendix
\onecolumn
\input{sec/appendix}
\end{document}

%% file: sec/intro.tex
\section{Introduction}

Diffusion models~\citep{ho2020DDPM,songDDIM,refitiedflow} have become a dominant paradigm for high-quality visual generation, but their iterative denoising process remains computationally expensive, especially for high-resolution image and video synthesis. This cost limits practical deployment and motivates a broad set of acceleration methods. Among them, feature caching~\citep{selvaraju2024fora,qiu2025acceleratingdiffusiontransformererroroptimized,zou2024DuCa,liu2024timestep,liu2025reusingforecastingacceleratingdiffusion,fengHiCache2025,zhengFoCa2025,Liu_2025} is particularly attractive because it is training-free: instead of running the full denoising network at every timestep, it reuses or predicts intermediate hidden states across nearby timesteps.

The key assumption behind feature caching is temporal redundancy. Hidden states at adjacent denoising steps are often similar, allowing expensive transformer blocks to be skipped with minor quality loss. However, this assumption becomes fragile under aggressive acceleration. As the cache interval grows, single-anchor reuse or forecasting accumulates hidden-state errors, which can lead to semantic drift, local artifacts, and degraded prompt fidelity. This creates a central trade-off: larger cache windows improve efficiency but increasingly deviate from the true denoising trajectory.

We start from a simple empirical observation. As shown in Figure~\ref{fig:layer-error}, cache-induced feature error tends to grow within each cache window, but drops sharply when a scheduled full-computation step is performed. This suggests that full-computation anchors partially pull the accelerated trajectory back toward the original denoising path. Rather than only using a full step as the starting point for the next cache window, we ask whether the newly obtained future anchor can also be used to correct the provisional states inside the previous active window.

Based on this observation, we propose \textit{Z-Cache}, a training-free and predictor-agnostic reflection framework for diffusion feature caching. For each active cache window, \textit{Z-Cache} first uses a base cache predictor to forecast intermediate hidden states from a past full-computation anchor and runs the corresponding provisional denoising transitions. After reaching the next scheduled full-computation anchor, \textit{Z-Cache} revisits the same window, estimates each intermediate hidden state with both the past and future anchors, and re-samples the window trajectory before committing the reflected window-end latent. Therefore, \textit{Z-Cache} is not a post-hoc smoothing operation over already generated samples. Instead, it is an in-loop trajectory correction mechanism that changes the subsequent denoising path. Extensive experiments demonstrate the effectiveness of Z-Cache in text-to-image, text-to-video, and image editing tasks, as well as its compatibility with other feature caching methods and other diffusion acceleration methods such as step distillation and quantization.

Our contributions are summarized as follows:
\begin{itemize}[leftmargin=*]
    \item We identify a cache-induced error pattern in diffusion feature caching: hidden-state errors tend to accumulate within a cache window but are sharply reduced after scheduled full-computation anchors, which motivates using future anchors to correct the previous active cache window.
    
    \item We propose \textit{Z-Cache}, a training-free cache-then-check framework that performs dual-anchor reflection and active-window re-sampling. Unlike post-hoc smoothing, \textit{Z-Cache} revises the latent trajectory inside the denoising loop and continues sampling from the reflected window-end state.
    
    \item We establish \textit{Z-Cache} as a plug-and-play and composable acceleration module. It can be attached to compatible feature-cache predictors, generalizes beyond free-form generation to image editing and video synthesis, and remains compatible with orthogonal acceleration techniques such as few-step distillation and low-bit quantization.
\end{itemize}

%% file: sec/related_work.tex
\section{Related Works}
\label{sec:related}

\paragraph{Feature caching}
Feature caching accelerates diffusion sampling by reusing or predicting intermediate activations. It spans U-Net models~\citep{li2023FasterDiffusion,ma2024deepcache}, DiT representation reuse~\citep{selvaraju2024fora,chen2024delta-dit}, adaptive timestep caching~\citep{liu2024timestep}, token-wise or dual feature caching~\citep{zou2024accelerating,zou2024DuCa}, and predictors such as TaylorSeer~\citep{liu2025reusingforecastingacceleratingdiffusion}, SpeCa~\citep{Liu_2025}, and MagCache~\citep{ma2025magcachefastvideogeneration}. SenCache~\citep{haghighi2026sencache} adapts reuse using sensitivity estimates, D2Cache~\citep{liu2026d2cache} improves temporal prediction with second-order deltas, and ResCa~\citep{fang2026resca} updates tokens using proxy-guided residuals.

\paragraph{Positioning of \textit{Z-Cache}}
\textit{Z-Cache} performs \textit{cache-then-check} correction: after observing the next scheduled full-computation anchor on the provisional trajectory, it combines past- and future-anchor predictions, replays the closed provisional window, and continues from the reflected endpoint. This complements compatible cache policies and predictors while preserving the full-computation schedule; composition requires compatible cached representations and state updates. Appendix~\ref{app:related_full} provides a detailed taxonomy and comparisons.

%% file: sec/method.tex
\section{Methodology}
\label{sec:method}

\begin{figure*}[t]
\vspace{-0.5cm}
    \centering
    \includegraphics[width=\linewidth]{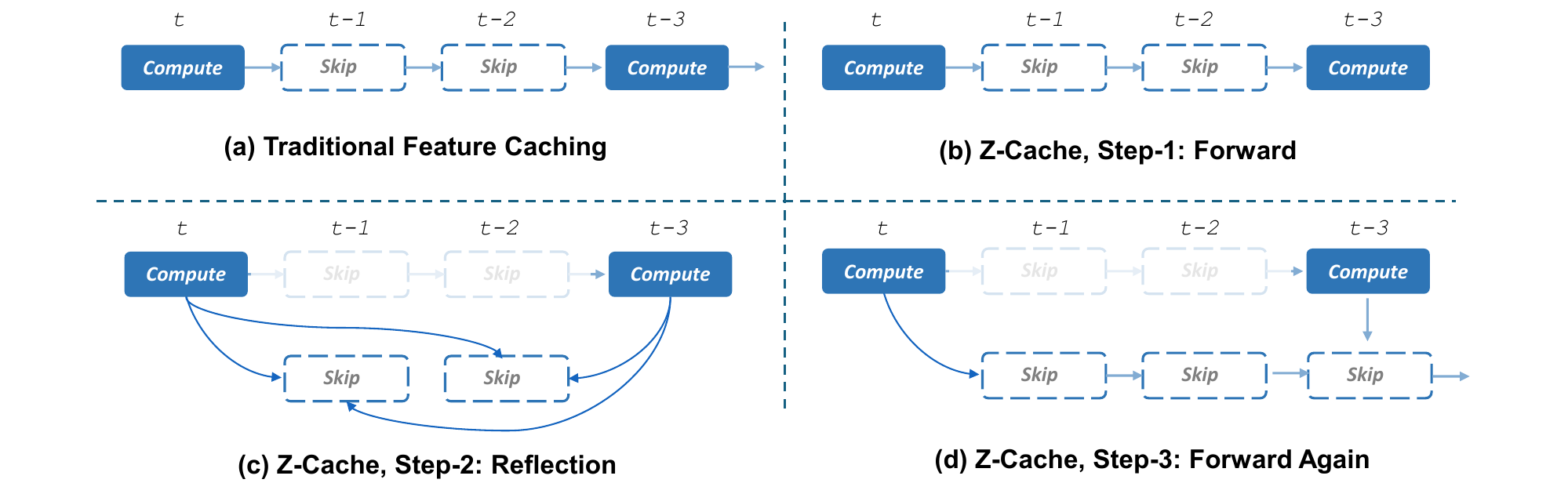}
    \captionsetup{width=\textwidth}
\caption{
\textbf{Pipeline of Z-Cache.}
The figure uses the reverse-diffusion notation $t,t-1,\ldots$ for visualization,
while the text uses a sampling-order index that increases along denoising.
Traditional feature caching forecasts an active window from a past full-computation
anchor and directly moves to the next window. In contrast, Z-Cache treats the
window as provisional: after obtaining the next scheduled full-computation anchor,
it revisits the same window, constructs dual-anchor hidden estimates, re-samples
the window trajectory, and commits the reflected window-end latent. Z-Cache
therefore performs in-loop trajectory correction without adding extra 
full-computation steps.
}
    \label{fig:pipelines}
\vspace{-0.3cm}
\end{figure*}

\subsection{Preliminary}
\label{sec:method_preliminary}

\subsubsection{Diffusion sampling.}
Diffusion models~\citep{ho2020DDPM,songDDIM} generate a clean sample by iteratively denoising a noisy latent.
In a standard DDPM-style parameterization, one reverse step is written as
\begin{equation}
\begin{aligned}
    p_\theta(x_{t-1}\mid x_t)
    =\mathcal{N}\!\left(
    x_{t-1};
    \frac{1}{\sqrt{\alpha_t}}\!\left(
    x_t-\frac{1-\alpha_t}{\sqrt{1-\bar{\alpha}_t}}\tau_\theta(x_t,t)
    \right),
    \beta_t\mathbf{I}
    \right),
\end{aligned}
\label{eq:ddpm_reverse}
\end{equation}
where $\tau_\theta$ is the denoising network and $\alpha_t,\bar\alpha_t,\beta_t$ are scheduler coefficients.
Modern samplers may use different update rules, but they share the same iterative structure.
To avoid confusion between the diffusion timestep $t$ and the order in which the sampler is executed, we use $i\in\{0,\ldots,S-1\}$ to denote the sampling-order index in the rest of this section, with larger $i$ corresponding to later denoising.
Let $\tau_i$ be the diffusion timestep used at sampling index $i$.
We write one denoising update abstractly as
\begin{equation}
    x_{i+1}=F_i(x_i,\mathcal{H}_i),
    \label{eq:sampler_update}
\end{equation}
where $\mathcal{H}_i$ denotes the hidden features produced by the denoising network at $x_i$ and $\tau_i$.

\subsubsection{Diffusion transformer hidden states.}
A Diffusion Transformer (DiT)~\citep{peebles2023dit} is composed of $L$ transformer blocks,
$\mathcal{G}=g_L\circ\cdots\circ g_1$, where each block contains attention and MLP submodules with timestep- and condition-dependent normalization.
At sampling index $i$, the full denoising network produces a collection of block features
\begin{equation}
    \mathcal{H}_i=\{h_i^\ell\}_{\ell=1}^{L}.
\end{equation}
Feature caching accelerates inference by avoiding full recomputation of $\mathcal{H}_i$ at selected sampling indices and replacing it with a cached or predicted estimate.

\subsection{Feature Caching}
\label{sec:feature_caching}

Let $\mathcal{A}_i$ be the set of sampling indices whose full hidden states have been cached before evaluating index $i$.
A generic cache predictor $\mathcal{C}$ estimates the hidden features at a cached step by
\begin{equation}
    \hat{\mathcal{H}}_i
    =
    \mathcal{C}\!\left(\{(j,\mathcal{H}_j):j\in\mathcal{A}_i\}, i\right).
    \label{eq:generic_cache_predictor}
\end{equation}
This abstraction covers simple feature reuse, timestep-adaptive caching, polynomial forecasting, speculative prediction, and calibration-style predictors.
For example, a naive fixed-interval cache method computes full features at an anchor $a$ and reuses them for the following cached steps,
\begin{equation}
    \hat{\mathcal{H}}_i=\mathcal{H}_a,
    \qquad a<i<b,
    \label{eq:naive_cache}
\end{equation}
where $[a,b]$ is one cache window and $b-a$ is the cache interval.
Larger intervals reduce the number of full denoising-network evaluations, but the approximation error
$\|\mathcal{H}_i-\hat{\mathcal{H}}_i\|$ can accumulate inside the window and lead to semantic drift or local artifacts.

\subsection{Z-Cache}
\label{sec:zcache}

\textit{Z-Cache} is a training-free reflection wrapper for feature caching.
It does not prescribe a specific cache predictor; instead, it can be applied on top of a base predictor $\mathcal{C}$ such as TaylorSeer, SpeCa, or MagCache.
The key idea is to treat the states inside the current cache window as provisional until the next scheduled full-computation anchor is available.
After obtaining that future anchor, \textit{Z-Cache} revisits the active window, constructs dual-anchor hidden estimates, re-samples the window trajectory, and then continues denoising from the reflected state.

\paragraph{Active cache window.}
Consider one active cache window $[a,b]$ in sampling-order indices.
Full hidden states are computed at the endpoint anchors $a$ and $b$, while the intermediate indices $a<i<b$ are cached.
Let $\mathcal{A}_a$ denote the anchor set available before entering this window, which includes the past anchor $a$.
The window proceeds in three stages.

\paragraph{Stage 1: forecast from the past anchor.}
Before the future endpoint is known, the base cache predictor estimates each intermediate hidden state from the currently available anchors:
\begin{equation}
    \hat{\mathcal{H}}^p_i
    =
    \mathcal{C}\!\left(\{(j,\mathcal{H}_j):j\in\mathcal{A}_a\}, i\right),
    \qquad a<i<b,
    \label{eq:past_predictor}
\end{equation}
where the superscript $p$ denotes the past-anchor prediction.
These estimates define a provisional trajectory,
\begin{equation}
    \bar{x}_a=x_a,
    \qquad
    \bar{x}_{i+1}=F_i(\bar{x}_i,\bar{\mathcal{H}}_i),
    \qquad a\le i<b,
    \label{eq:forecast_trajectory}
\end{equation}
where $\bar{\mathcal{H}}_a=\mathcal{H}_a$ and $\bar{\mathcal{H}}_i=\hat{\mathcal{H}}^p_i$ for $a<i<b$.
After the provisional trajectory reaches the endpoint, the sampler performs the next scheduled full computation and obtains the future anchor $\mathcal{H}_b$.
This full computation is part of the original cache schedule, so \textit{Z-Cache} does not add extra scheduled full-computation steps.

\paragraph{Stage 2: dual-anchor reflection.}
Once $\mathcal{H}_b$ is available, \textit{Z-Cache} constructs a future-anchor prediction for each intermediate index by reusing the same base predictor with the extended anchor set:
\begin{equation}
    \hat{\mathcal{H}}^f_i
    =
    \mathcal{C}\!\left(\{(j,\mathcal{H}_j):j\in\mathcal{A}_a\cup\{b\}\}, i\right),
    \qquad a<i<b,
    \label{eq:future_predictor}
\end{equation}
where the superscript $f$ denotes the future-anchor prediction.
The reflected hidden state is the convex dual-anchor estimate
\begin{equation}
    \tilde{\mathcal{H}}_i
    =
    (1-w_i)\hat{\mathcal{H}}^p_i
    +
    w_i\hat{\mathcal{H}}^f_i,
    \qquad a<i<b.
    \label{eq:reflected_hidden}
\end{equation}
The weight $w_i$ is defined by the normalized progress inside the active window,
\begin{equation}
    w_i=\frac{i-a}{b-a},
    \qquad w_i\in[0,1],
    \label{eq:window_progress}
\end{equation}

This schedule is hyperparameter-free once the cache interval is fixed.
It trusts the past-anchor prediction near the beginning of the window and gradually shifts trust toward the future-anchor prediction near the endpoint.

\paragraph{Stage 3: active-window re-sampling.}
The reflection step is performed inside the denoising loop, not after a sample has already been finalized.
The provisional latent states $\bar{x}_{a+1},\ldots,\bar{x}_b$ are discarded, and the active window is re-sampled with the reflected hidden states:
\begin{equation}
    \tilde{x}_a=x_a,
    \qquad
    \tilde{x}_{i+1}=F_i(\tilde{x}_i,\tilde{\mathcal{H}}_i),
    \qquad a\le i<b,
    \label{eq:resample_trajectory}
\end{equation}
where $\tilde{\mathcal{H}}_a=\mathcal{H}_a$ and $\tilde{\mathcal{H}}_i$ is given by Eq.~\eqref{eq:reflected_hidden} for $a<i<b$. The transition at $a$ is reused.
After re-sampling, \textit{Z-Cache} commits $\tilde{x}_{b+1}=F_b(\tilde{x}_b,\mathcal H_b)$ using the retained right-anchor features and updates the cache accordingly (Appendix~\ref{app:zcache_pseudocode}).
States before $a$ have already been committed and are not modified, while states after $b+1$ have not yet been generated.
Thus, \textit{Z-Cache} changes the subsequent denoising trajectory without creating inconsistency with finalized states outside the current window.
This distinguishes \textit{Z-Cache} from post-hoc hidden-state smoothing: the reflected hidden estimates affect later outputs because they are used to re-run the latent transitions inside the active window.

\paragraph{Local error propagation.}
We next give a local analysis explaining why active-window reflection can reduce trajectory error.
Let the full, non-accelerated trajectory satisfy
\begin{equation}
    x^\star_{i+1}=F_i(x^\star_i,\mathcal{H}^\star_i),
    \label{eq:full_trajectory}
\end{equation}
and let an accelerated method use a hidden estimate $\bar{\mathcal{H}}_i$ and latent $\bar{x}_i$.
Define the latent error and hidden approximation error as
\begin{equation}
    \delta_i=\|\bar{x}_i-x^\star_i\|,
    \qquad
    \epsilon_i=\|\bar{\mathcal{H}}_i-\mathcal{H}^\star_i\|.
\end{equation}
Assume each local update is Lipschitz in both the latent state and the hidden state:
\begin{equation}
    \|F_i(x,\mathcal{H})-F_i(x',\mathcal{H}')\|
    \le
    L_i^x\|x-x'\|+L_i^H\|\mathcal{H}-\mathcal{H}'\|.
    \label{eq:local_lipschitz}
\end{equation}
Then
\begin{equation}
    \delta_{i+1}\le L_i^x\delta_i+L_i^H\epsilon_i.
\end{equation}
Unrolling this recursion over one active window gives
\begin{equation}
    \delta_b
    \le
    \left(\prod_{i=a}^{b-1}L_i^x\right)\delta_a
    +
    \sum_{i=a}^{b-1}
    \left(\prod_{j=i+1}^{b-1}L_j^x\right)L_i^H\epsilon_i,
    \label{eq:latent_error_bound}
\end{equation}
where an empty product equals $1$.
Therefore, the error $\delta_b$ depends on the hidden-state errors accumulated at all earlier indices in the same window.
This explains why correcting intermediate cached steps changes later latents: the sampler recursively propagates hidden-state errors through Eq.~\eqref{eq:sampler_update}.

For the no-reflection base predictor, the hidden error at an intermediate index is
\begin{equation}
    \epsilon_i^p=\|\hat{\mathcal{H}}^p_i-\mathcal{H}^\star_i\|.
\end{equation}
For \textit{Z-Cache}, Eq.~\eqref{eq:reflected_hidden} and the triangle inequality yield
\begin{equation}
\begin{aligned}
    \epsilon_i^z
    &=
    \|\tilde{\mathcal{H}}_i-\mathcal{H}^\star_i\| \\
    &\le
    (1-w_i)\epsilon_i^p+w_i\epsilon_i^f,
\end{aligned}
\label{eq:reflected_hidden_bound}
\end{equation}
where $\epsilon_i^f=\|\hat{\mathcal{H}}^f_i-\mathcal{H}^\star_i\|$ is the future-anchor prediction error.
Combining Eq.~\eqref{eq:reflected_hidden_bound} with Eq.~\eqref{eq:latent_error_bound} shows that no-reflection caching is the special case $w_i=0$, while \textit{Z-Cache} replaces the single-anchor error term $\epsilon_i^p$ with the reflected error term $\epsilon_i^z$.
Thus, whenever the dual-anchor estimate reduces hidden error over the active window, \textit{Z-Cache} gives a tighter local upper bound on $\delta_b$.

%% file: sec/experiments.tex
\section{Experiments}

\subsection{Class-to-Image Generation}

For comprehensive evaluation of class-conditional generation on ImageNet \citep{Imagenet}, all experiments employed a DDIM-50 sampling protocol at $256\times256$ resolution with 50,000 images generated in total. Methods implementing activation period optimization are explicitly annotated with the operator $\mathcal{N}$. All experiments were executed on NVIDIA H20 GPUs with standardized controls.

\input{tab/DiT-Metrics}

Quantitative evaluations on ImageNet class-to-image generation using the DiT-XL/2 model confirm that \textit{Z-Cache} consistently outperforms state-of-the-art caching baselines across key metrics. As Table~\ref{table:DiT_Metrics} demonstrates, \textit{Z-Cache} achieves the best performance at all tested acceleration levels. Specifically, at $6.22\times$ acceleration, \textit{Z-Cache} maintains both the lowest FID (3.25) and sFID (5.52) score, and also the highest Inception Score (219.30). On the contrary, FORA, ToCa and DuCa are all suffering from catastrophic degradation (FID $> 10$, sFID $> 10$, and Inception Score $<200$). This signifies \textit{Z-Cache}'s superior preservation of image fidelity and diversity.

Furthermore, as detailed in Table~\ref{table:DiT_Metrics}, \textit{Z-Cache} achieved a comparable performance with Taylorseer while surpassing other sampling results on the $6.22\times$ acceleration level. This comprehensive performance advantage underscores that \textit{Z-Cache} generates sampling trajectories closely aligned with the original model, minimizing semantic shifts in the output data.

\begin{figure}[t]
        \vspace{-0.25cm}\centering\includegraphics[width=\linewidth]
{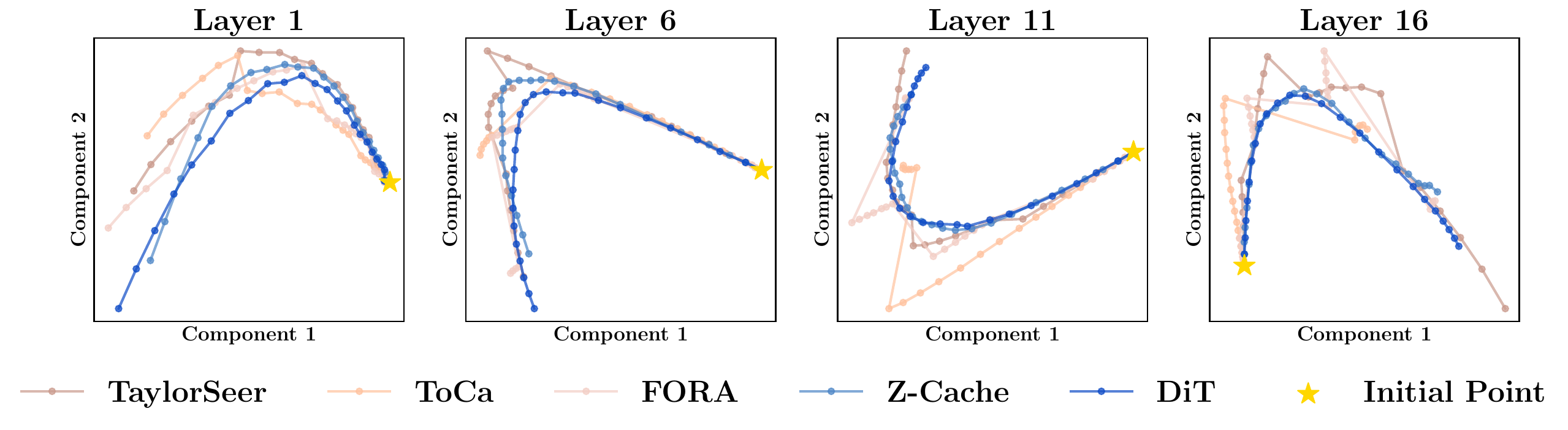}
    \captionsetup{width=\linewidth}
    \vspace{-0.6cm}
    \caption{The Trajectories of PCA components of hidden features. \textit{Z-Cache} has the minimum divergence to the ground-truth comparing with other methods.}
    \vspace{-0.3cm}
    \label{fig:pca-features}
\end{figure}

To better illustrate \textit{Z-Cache}'s mechanism, we gathered latent hidden states generated during sampling and compared their trajectories. Visualized in Figure \ref{fig:pca-features}, other methods often produce "bumps" during cache time as they do not perceive future states, while \textit{Z-Cache} repetitively reflects past hidden states and therefore lies close to the ground truth trajectories.

\subsection{Text-to-Image Generation}
\label{sec:t2i}
\input{tab/t2i}

The paired comparisons in Table~\ref{tab:full_paired_base_predictors}
show a consistent pattern: \textit{Z-Cache} mainly acts as a quality-recovery
module for aggressive feature caching. Across TeaCache, TaylorSeer,
SpeCa, and MagCache, reflection improves PSNR, SSIM, and LPIPS in
every matched setting, while ImageReward improves in all but one case. These results indicate \textit{Z-Cache} can be directly applied to any existing feature caching methods.

Table~\ref{tab:full_paired_base_predictors} also separates arithmetic cost from
practical wall-clock cost. Since \textit{Z-Cache} reuses the next scheduled
full-computation anchor rather than inserting extra full steps, the additional
FLOPs are small. The FLOPs overhead is below $0.7\%$ for all paired settings and
below $0.2\%$ for TaylorSeer, SpeCa, and MagCache, so the FLOPs speedup is
nearly unchanged after reflection. However, latency and peak memory are more
predictor-dependent. For TaylorSeer and SpeCa, \textit{Z-Cache} introduces modest latency
overhead and no measured increase in peak memory. For TeaCache and MagCache,
the quality gain remains consistent, but the current implementation incurs a
larger wall-clock cost and higher peak memory. This suggests that \textit{Z-Cache} should
be viewed as a FLOPs-light quality-recovery wrapper rather than a universally
latency-free module.

\begin{figure}[h]
    \vspace{-0.3cm}
    \centering
    \includegraphics[width=\linewidth]{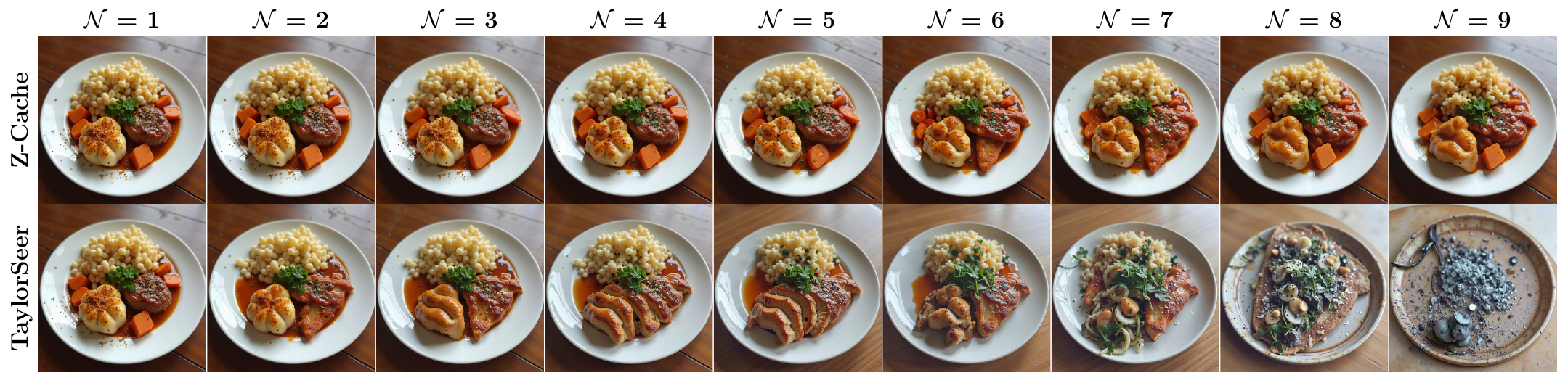}
    \captionsetup{width=\linewidth}
    \caption{\textbf{Semantic stability under increasing cache interval $\mathcal{N}$.}
    As the cache interval increases, TaylorSeer gradually accumulates semantic drift,
    while Z-Cache maintains more stable object layout and visual identity.}
    \label{fig:seq-n}
    \vspace{-0.2cm}
\end{figure}

Figure~\ref{fig:seq-n} further shows an interval sweep where TaylorSeer
progressively loses semantic details as $\mathcal{N}$ increases, while \textit{Z-Cache}
preserves the object layout and visual identity over larger cache intervals.
Together, the prompt-level comparison and the interval sweep are consistent with
the quantitative trend in Table~\ref{tab:full_paired_base_predictors}: revisiting
the active cache window with a future anchor reduces cache-induced semantic drift
under high acceleration.

\subsection{Text-to-Video Generation}
\input{tab/HunyuanVideo-Metrics}

\begin{figure*}[h] 
    \centering
    \includegraphics[width=\linewidth]{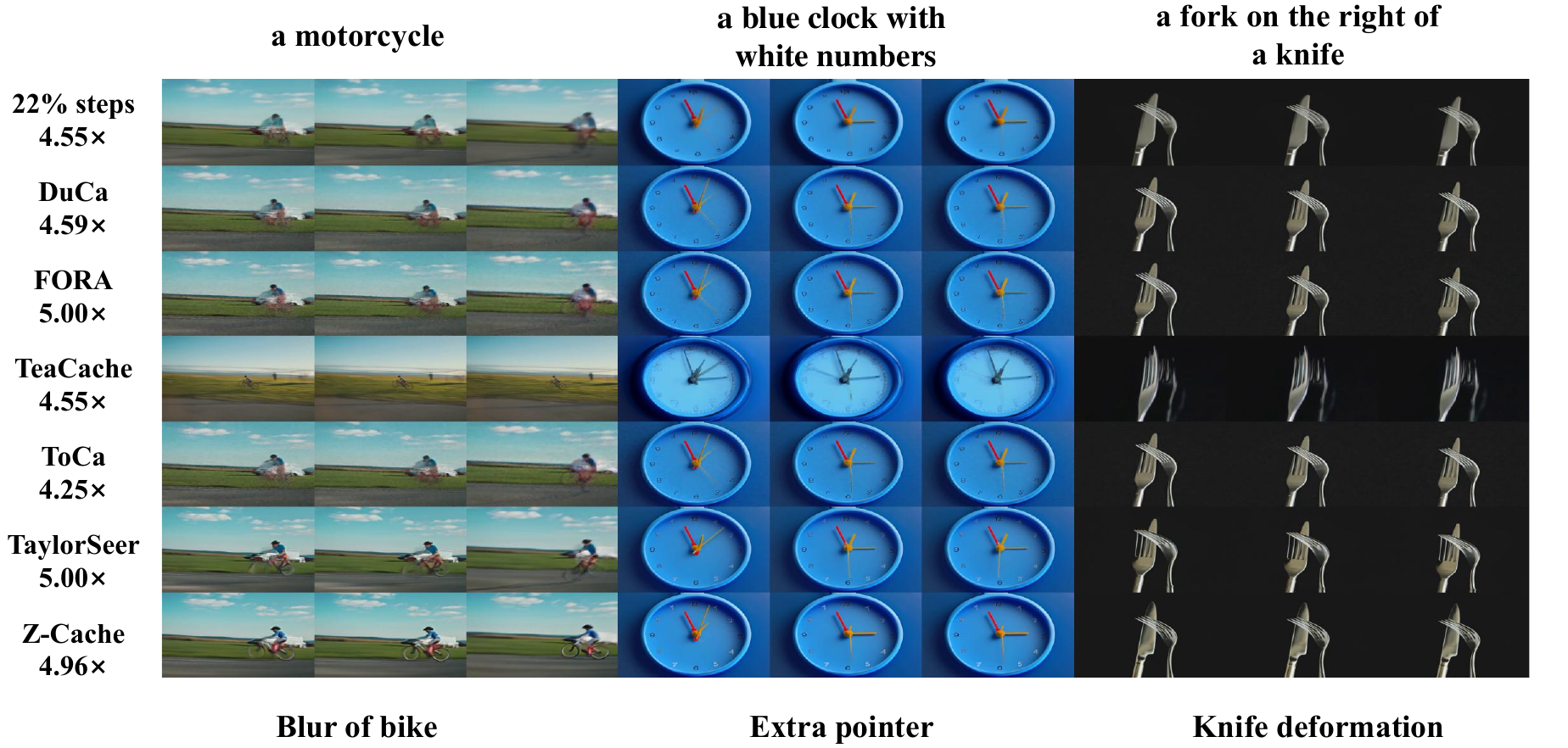}
    \captionsetup{width=\linewidth}
    \caption{Visualization of videos generated by Hunyuan Video with different feature caching methods. The annotated speedups denote FLOPs reductions relative to 50-step sampling.}
    \label{fig:vbench}
\end{figure*}
Extending to video synthesis on HunyuanVideo, \textit{Z-Cache} maintains a consistent and comparable performance with other competing methods on the VBench metric. As Table~\ref{table:HunyuanVideo-Metrics} shows, our method achieves the highest VBench scores within the two caching groups (80.10 at 4.96$\times$ FLOPs speedup and 79.60 at 6.20$\times$ FLOPs speedup). This demonstrates \textit{Z-Cache}'s advantage for video diffusion acceleration, where existing caching strategies struggle to balance efficiency with temporal consistency.

Beyond quantitative metrics, we present qualitative comparisons of these methods on identical frame sequences in Figure \ref{fig:vbench}. For instance, in the “a motorcycle” video, \textit{Z-Cache} generates a coherent depiction of the rider’s face, whereas methods like TaylorSeer produce distorted facial features or fragmented figures. In the ``a blue clock with white numbers'' video, several baselines generate extra or distorted clock hands, whereas Z-Cache better preserves the intended clock structure. Finally, in the \textit{“a fork on the right of a knife”} video, all baseline methods either produce malformed or misshapen knives or completely disregard the fork, whereas \textit{Z-Cache} accurately captures both objects in their correct spatial arrangement.

\subsection{Miscellaneous}

To test whether \textit{Z-Cache} remains effective in the few-step regime, we additionally evaluate it on a distilled model, as reported in Table~\ref{tab:fewstep}. Specifically, we compare \textit{Z-Cache} with TaylorSeer and ToCa on Qwen-Image-Lightning using an 8-step sampler. \textit{Z-Cache} achieves the highest PSNR and the lowest LPIPS among the methods, while maintaining comparable SSIM. These results show that, even under aggressive distillation with very few denoising steps, \textit{Z-Cache} still delivers a clear improvement over prior cache-based baselines, indicating that cache-based acceleration can remain beneficial beyond standard multi-step diffusion settings.

\begin{table*}[h]
\centering
\caption{
Additional quantitative results on few-step generation, image editing,
U-Net backbones, and non-cache acceleration settings.  All \textbf{Z-Cache} variants use the TaylorSeer predictor, and the number in parentheses is the cache interval.
}
\label{tab:additional_quantitative_results}
\setlength{\tabcolsep}{4pt}
\renewcommand{\arraystretch}{1.05}

\begin{subtable}[t]{0.38\textwidth}
\centering
\caption{Few-step distilled model Qwen-Image-Lightning on DrawBench.}
\label{tab:fewstep}
\scriptsize
\resizebox{\linewidth}{!}{
\begin{tabular}{lccc}
\toprule
Method & PSNR $\uparrow$ & SSIM $\uparrow$ & LPIPS $\downarrow$ \\
\midrule
TaylorSeer (2) & 30.222 & 0.6734 & 0.2984 \\
ToCa (2) & 29.546 & 0.6780 & 0.2938 \\
\rowcolor{gray!15}
\textbf{Z-Cache (2)}
& \textbf{30.440} & \textbf{0.6785} & \textbf{0.2910} \\
\bottomrule
\end{tabular}
}
\end{subtable}
\hfill
\begin{subtable}[t]{0.58\textwidth}
\centering
\caption{Comparison with other acceleration methods on DrawBench.}
\label{tab:flux-non-cache}
\scriptsize
\resizebox{\linewidth}{!}{
\begin{tabular}{lccccc}
\toprule
Method
& PSNR $\uparrow$
& SSIM $\uparrow$
& LPIPS $\downarrow$
& ImageReward $\uparrow$
& Accel. $\uparrow$ \\
\midrule
ToMA~\citep{lu2025tomatokenmergeattention}
& 29.271 & 0.6984 & 0.4258 & 0.7812 & 1.21$\times$ \\
8bit
& 32.331 & \textbf{0.8334} & 0.1770 & \textbf{1.1971} & 2.10$\times$ \\
8bit TaylorSeer (2)
& 33.085 & 0.8197 & 0.1413 & 0.8773 & 3.12$\times$ \\
\rowcolor{gray!15}
\textbf{8bit Z-Cache (2)}
& \textbf{33.600} & 0.8289 & \textbf{0.1337} & 0.8877 & 3.12$\times$ \\
Flux-mini
& 28.101 & 0.5186 & 0.6449 & 0.2522 & \textbf{3.80$\times$} \\
\rowcolor{gray!15}
\textbf{Z-Cache (4)}
& 30.570 & 0.7758 & 0.2406 & 0.9890 & 2.17$\times$ \\
\bottomrule
\end{tabular}
}
\end{subtable}

\vspace{0.9em}

\begin{subtable}[t]{0.38\textwidth}
\centering
\caption{FLUX-Kontext-dev~\citep{labs2025flux1kontextflowmatching} on Gedit-bench~\citep{liu2025step1xeditpracticalframeworkgeneral}.}
\label{tab:flux-edit}
\scriptsize
\resizebox{\linewidth}{!}{
\begin{tabular}{lccc}
\toprule
Method & PSNR $\uparrow$ & SSIM $\uparrow$ & LPIPS $\downarrow$ \\
\midrule
TaylorSeer (8) & 29.014 & 0.4338 & 0.4448 \\
\rowcolor{gray!15}
\textbf{Z-Cache (8)}
& \textbf{29.511} & \textbf{0.5114} & \textbf{0.3694} \\
\midrule
TaylorSeer (7) & 29.675 & 0.5013 & 0.3658 \\
\rowcolor{gray!15}
\textbf{Z-Cache (7)}
& \textbf{30.279} & \textbf{0.5667} & \textbf{0.3015} \\
\bottomrule
\end{tabular}
}
\end{subtable}
\hfill
\begin{subtable}[t]{0.58\textwidth}
\centering
\caption{Experiments with the U-Net model SDXL~\citep{podell2023sdxlimprovinglatentdiffusion} on DrawBench.}
\label{tab:unet}
\scriptsize
\resizebox{\linewidth}{!}{
\begin{tabular}{lccccc}
\toprule
Method
& PSNR $\uparrow$
& SSIM $\uparrow$
& LPIPS $\downarrow$
& ImageReward $\uparrow$
& Accel. $\uparrow$ \\
\midrule
TaylorSeer (5)
& 29.028 & 0.6969 & 0.3294 & 0.3350 & 2.61$\times$ \\
\rowcolor{gray!15}
\textbf{Z-Cache (5)}
& \textbf{29.399} & \textbf{0.7112} & \textbf{0.3100}
& \textbf{0.3788} & 2.61$\times$ \\
\midrule
TaylorSeer (7)
& 28.251 & 0.6041 & 0.4528 & 0.1024 & 4.00$\times$ \\
\rowcolor{gray!15}
\textbf{Z-Cache (7)}
& \textbf{28.589} & \textbf{0.6373} & \textbf{0.4162}
& \textbf{0.1802} & 4.00$\times$ \\
\bottomrule
\end{tabular}
}
\end{subtable}

\end{table*}

In addition to free-form generation, we also evaluate \textit{Z-Cache} in an image-editing setting. Specifically, we run FLUX-Kontext-dev on the Gedit-bench benchmark and compare \textit{Z-Cache} with TaylorSeer, as summarized in Table~\ref{tab:flux-edit}. At the same acceleration ratio, \textit{Z-Cache} achieves higher PSNR and SSIM while substantially reducing LPIPS. These improvements indicate that \textit{Z-Cache} better preserves the source image’s structure and perceptual details, which are crucial for maintaining subject identity and keeping non-edited background regions consistent. 

We further compare \textit{Z-Cache} with non-caching acceleration methods in Table~\ref{tab:flux-non-cache}. The results show that \textit{Z-Cache} is complementary to quantization: adding \textit{Z-Cache} on top of 8-bit quantization (8bit \textit{Z-Cache}) increases the speedup from 2.10$\times$ (8bit only) to 3.12$\times$, while also improving PSNR and reducing LPIPS. Under the same 3.12$\times$ acceleration budget, 8bit \textit{Z-Cache} consistently outperforms 8bit TaylorSeer across all four metrics (PSNR, SSIM, LPIPS, and ImageReward), indicating that the reflection step provides non-trivial gains even when the backbone is already compressed. In contrast, a distilled model such as Flux-mini achieves a comparable speedup (3.80$\times$) but with substantially worse fidelity and preference scores, suggesting that cache-based acceleration remains competitive in the high-speed regime. Overall, these results support the view that Z-Cache is not a replacement for quantization/pruning, but a plug-in that can be layered on top of them.

We additionally evaluate \textit{Z-Cache} on \textbf{SDXL}, a U-Net-based diffusion model, and directly compare it with TaylorSeer under matched step budgets (i.e., the same cache-step setting). As summarized in Table~\ref{tab:unet}, \textit{Z-Cache} consistently improves PSNR and SSIM, reduces LPIPS, and achieves higher ImageReward than TaylorSeer across both acceleration ratios, while using the same number of cache steps. These results indicate that the proposed cache-then-check mechanism generalizes beyond DiT-style backbones and naturally transfers to U-Net architectures.

Additional trajectory controls, fusion and replay ablations, and recent-predictor
compositions are reported in
Appendix~\ref{app:supplementary_experiments}.

%% file: tab/DiT-Metrics.tex
\begin{table*}[t]
\vspace{-0.2cm}
\centering
\caption{\textbf{Quantitative comparison on class-to-image generation} on ImageNet with DiT-XL/2. F-Speed denotes the FLOPs reduction relative to DDIM with 50 steps. Metrics marked with ``-'' are left unreported, as they were not provided in the original publications.}
    \vspace{-0.3cm}
\setlength\tabcolsep{7pt} 
\small
\resizebox{\textwidth}{!}{\begin{tabular}{l | c | c c c | c c |c|ccc}
\toprule
\bf Method  & \makecell{\bf Efficient \\ \bf Attention} & \bf Latency(s) $\downarrow$ & \bf FLOPs(T) $\downarrow$ & \bf F-Speed $\uparrow$  & \bf FID $\downarrow$ & \bf sFID $\downarrow$ &   \makecell{\bf Inception\\ \bf Score} $\uparrow$  & \bf PSNR $\uparrow$ & \bf SSIM $\uparrow$ & \bf   LPIPS $\downarrow$ \\
\midrule
{\textbf{$\text{DDIM-50 steps}$}} & \ding{52}& {0.428}  & {23.74}  & {1.00$\times$}  &  {{2.32}} &  {{4.32}}   &{241.25}&$\infty$&1.00&0.000\\
\midrule

{\textbf{$\text{DDIM-12 steps}$}}  & \ding{52}& {0.128}  & {5.70}  & {4.17$\times$}  &  {{7.80}} &  {{8.03}} &{184.50}&22.70&0.742&0.196\\

\textbf{L2C(NFE=30)}&\ding{52}&0.281&11.55&2.05$\times$&2.61&-&237.83&-&-&-\\

\textbf{SmoothCache}($\alpha=0.22$)&\ding{52}&0.251&8.57&2.77$\times$&4.15&-&231.71&-&-&-\\

{\textbf{$\text{DDIM-10 steps}$}}  & \ding{52}& {0.115}  & {4.75}  & {5.00$\times$}  &  {{12.15}} &  {{11.33}} &{159.13}&21.52&0.687&0.251\\

\textbf{FORA} ($\mathcal{N}=5$) & \ding{52} & 0.149 & 5.24 & {4.53$\times$}  & 6.58 & 11.29  &{193.01}& 21.85&0.692&0.235\\

\textbf{\texttt{ToCa}} ($\mathcal{N}=6$)   & \ding{56}& {0.163} & {6.34} & {3.75$\times$} & {6.55}   & {7.10} &{189.53}& 17.66&0.572&0.360\\

\textbf{\texttt{DuCa}} ($\mathcal{N}=5$)  & \ding{52}& {0.154} & {6.27} & {3.78$\times$} & {6.06}  & 6.72  &{198.46} & 16.65&0.543&0.390\\

$\textbf{TaylorSeer} $ $(\mathcal{N}=5)$& \ding{52}& {0.195} & {5.24} & {4.53$\times$} & {2.65}  & {5.36}  &{231.59}& 28.50  & 0.901  &0.065\\

\rowcolor{gray!20}
$\textbf{Z-Cache} $ $(\mathcal{N}=5)$& \ding{52}& {0.213} & {5.25} & {4.53$\times$} & {\textbf{2.46}}  & {\textbf{4.71}}  &{\textbf{232.34}} &{\textbf{29.24}}  & {\textbf{0.908}}  &{\textbf{0.052}}\\

\midrule 

\textbf{FORA} ($\mathcal{N}=6$) & \ding{52} & 0.135 & 4.29 & {5.53$\times$}  & 9.24 & 14.84  &{171.33}& 20.69&0.638&0.290\\

\textbf{\texttt{ToCa}} ($\mathcal{N}=9$)   & \ding{56}& {0.127} & {4.54} & {5.23$\times$} & {12.86}   & {12.82} &{151.37}& 15.71   & 0.485 &0.473\\

\textbf{\texttt{DuCa}} ($\mathcal{N}=9$)  & \ding{52}& {0.105} & {4.30} & {5.52$\times$} & {12.05}  & 11.82  &{156.20}& 15.85  & 0.483  &0.460\\

$\textbf{TaylorSeer} $ $(\mathcal{N}=6)$& \ding{52}&{ 0.191 } & {4.76} & {4.98$\times$} & {3.09}  & {6.50}  &{225.16}& 25.48  & 0.831  &0.107\\

\rowcolor{gray!20}
$\textbf{Z-Cache} $ $(\mathcal{N}=6)$& \ding{52}& {0.207} & {4.78} & {4.96$\times$} & {\textbf{2.68}}  & {\textbf{4.84}}  &{\textbf{228.26}}& {\textbf{27.16}}  & {\textbf{0.880}}  &{\textbf{0.078}}\\

\midrule 

\textbf{FORA} ($\mathcal{N}=7$) & \ding{52} & 0.123 & 3.82 & {6.22$\times$}  & 12.55 & 18.63  &148.44& 19.87&0.598&0.335\\

\textbf{\texttt{ToCa}} ($\mathcal{N}=13$)   & \ding{56}& { 0.118 } & {4.03} & {5.90$\times$} & {21.24}   & {19.93} &{116.08}& 15.06   & 0.449 &0.540\\

\textbf{\texttt{DuCa}} ($\mathcal{N}=12$)  & \ding{52}& {0.081} & {3.94} & {6.02$\times$} & {31.97}  & 27.26  &{87.94} & 14.23 & 0.396  &0.593\\

$\textbf{TaylorSeer} $ $(\mathcal{N}=7)$& \ding{52}& { 0.177 } & {3.82} & {6.22$\times$} & {3.60}  & {7.07}  &{217.79}& \textbf{24.09}  & 0.794  &0.136\\

\rowcolor{gray!20}
$\textbf{Z-Cache} $ $(\mathcal{N}=7)$& \ding{52}& {0.185} & {3.82} & {6.22$\times$} & {\textbf{3.25}}  & {\textbf{5.52}}  &{\textbf{219.30}}& {23.96}  & {\textbf{0.795}}  &{\textbf{0.129}}\\

\rowcolor{gray!20}
\bottomrule
\end{tabular}}
\label{table:DiT_Metrics}
\end{table*}

%% file: tab/t2i.tex

\begin{table*}[h]
\centering
\caption{
\textbf{Full paired results for cache predictors on FLUX.1-dev with DrawBench.}
Rows are grouped by the equivalent cache interval $\mathcal{N}$.
Bold quality metrics indicate the better value within each Base/$+$Z-Cache pair; gray rows identify $+$Z-Cache.
Parameters are abbreviated in parentheses: TeaCache uses $(L)$, where $L=\texttt{rel\_l1\_thresh}$;
SpeCa uses $(B,D,m,M)=(\texttt{base},\texttt{decay},\texttt{min},\texttt{max})$;
MagCache uses $(T,K,R)=(\texttt{th},K,\texttt{retention})$.
}
\setlength\tabcolsep{4pt}
\small
\resizebox{\textwidth}{!}{
\begin{tabular}{l | c | c c | c c | c c c c}
\toprule
\rowcolor{white}
\makecell{\textbf{Method}\\\textbf{(FLUX.1-dev)}} &
\makecell{\textbf{Peak Mem}\\\textbf{(GiB)} $\downarrow$} &
\textbf{Latency(s)} $\downarrow$ &
\textbf{L-Speed} $\uparrow$ &
\textbf{FLOPs(T)} $\downarrow$ &
\textbf{F-Speed} $\uparrow$ &
\textbf{PSNR} $\uparrow$ &
\textbf{SSIM} $\uparrow$ &
\textbf{LPIPS} $\downarrow$ &
\textbf{IR} $\uparrow$ \\
\midrule

\multicolumn{10}{l}{\textbf{Equivalent cache interval $\mathcal{N}=6$}} \\
\midrule
\textbf{TeaCache} $(L=.8)$
& 35.77 & 2.53 & 3.67$\times$ & 891.97 & 4.17$\times$
& 28.47 & 0.6084 & 0.4691 & \textbf{1.0246} \\
\rowcolor{gray!20}
\textbf{TeaCache + Z-Cache} $(L=.8)$
& 39.99 & 5.07 & 1.83$\times$ & 896.71 & 4.15$\times$
& \textbf{28.65} & \textbf{0.6367} & \textbf{0.4328} & 1.0147 \\

\textbf{TaylorSeer}
& 39.96 & 3.10 & 3.00$\times$ & 745.54 & 4.99$\times$
& 28.84 & 0.6302 & 0.4359 & 0.9737 \\
\rowcolor{gray!20}
\textbf{TaylorSeer + Z-Cache}
& 39.96 & 3.29 & 2.82$\times$ & 746.47 & 4.98$\times$
& \textbf{29.22} & \textbf{0.6539} & \textbf{0.3914} & \textbf{0.9882} \\

\textbf{SpeCa} $(100,1,4,20)$
& 40.02 & 3.24 & 2.86$\times$ & 756.87 & 4.91$\times$
& 28.78 & 0.6617 & 0.4034 & 0.9517 \\
\rowcolor{gray!20}
\textbf{SpeCa + Z-Cache} $(100,1,4,20)$
& 40.02 & 3.46 & 2.69$\times$ & 757.96 & 4.91$\times$
& \textbf{29.18} & \textbf{0.6806} & \textbf{0.3675} & \textbf{0.9844} \\

\textbf{MagCache} $(.45,7,.02)$
& 33.85 & 1.99 & 4.67$\times$ & 746.03 & 4.99$\times$
& 29.44 & 0.7119 & 0.3556 & 0.9316 \\
\rowcolor{gray!20}
\textbf{MagCache + Z-Cache} $(.45,7,.02)$
& 37.98 & 3.37 & 2.75$\times$ & 747.27 & 4.98$\times$
& \textbf{29.73} & \textbf{0.7253} & \textbf{0.3236} & \textbf{0.9674} \\

\midrule
\multicolumn{10}{l}{\textbf{Equivalent cache interval $\mathcal{N}=7$}} \\
\midrule
\textbf{TeaCache} $(L=1.2)$
& 35.77 & 1.99 & 4.67$\times$ & 668.98 & 5.56$\times$
& 28.30 & 0.5577 & 0.5466 & 0.8715 \\
\rowcolor{gray!20}
\textbf{TeaCache + Z-Cache} $(L=1.2)$
& 39.99 & 4.75 & 1.95$\times$ & 673.17 & 5.53$\times$
& \textbf{28.41} & \textbf{0.5875} & \textbf{0.5085} & \textbf{0.8802} \\

\textbf{TaylorSeer}
& 39.96 & 2.94 & 3.16$\times$ & 671.11 & 5.54$\times$
& 28.61 & 0.6032 & 0.4797 & 0.9455 \\
\rowcolor{gray!20}
\textbf{TaylorSeer + Z-Cache}
& 39.96 & 3.12 & 2.97$\times$ & 672.02 & 5.53$\times$
& \textbf{28.93} & \textbf{0.6310} & \textbf{0.4306} & \textbf{0.9753} \\

\textbf{SpeCa} $(100,1,5,20)$
& 40.02 & 2.97 & 3.12$\times$ & 679.74 & 5.47$\times$
& 28.54 & 0.6353 & 0.4527 & 0.9305 \\
\rowcolor{gray!20}
\textbf{SpeCa + Z-Cache} $(100,1,5,20)$
& 40.02 & 3.39 & 2.74$\times$ & 680.83 & 5.46$\times$
& \textbf{28.87} & \textbf{0.6581} & \textbf{0.4059} & \textbf{0.9698} \\

\textbf{MagCache} $(.45,9,.02)$
& 33.85 & 1.94 & 4.78$\times$ & 671.50 & 5.54$\times$
& 29.41 & 0.7020 & 0.3689 & 0.9252 \\
\rowcolor{gray!20}
\textbf{MagCache + Z-Cache} $(.45,9,.02)$
& 37.98 & 3.12 & 2.97$\times$ & 672.70 & 5.53$\times$
& \textbf{29.68} & \textbf{0.7164} & \textbf{0.3356} & \textbf{0.9682} \\

\midrule
\multicolumn{10}{l}{\textbf{Equivalent cache interval $\mathcal{N}=8$}} \\
\midrule
\textbf{TeaCache} $(L=1.4)$
& 35.77 & 1.87 & 4.96$\times$ & 594.65 & 6.25$\times$
& 28.24 & 0.5353 & 0.5770 & 0.7680 \\
\rowcolor{gray!20}
\textbf{TeaCache + Z-Cache} $(L=1.4)$
& 39.99 & 4.64 & 2.00$\times$ & 598.66 & 6.21$\times$
& \textbf{28.33} & \textbf{0.5702} & \textbf{0.5321} & \textbf{0.8288} \\

\textbf{TaylorSeer}
& 39.96 & 2.83 & 3.28$\times$ & 596.69 & 6.23$\times$
& 28.46 & 0.5703 & 0.5190 & 0.9292 \\
\rowcolor{gray!20}
\textbf{TaylorSeer + Z-Cache}
& 39.96 & 3.17 & 2.93$\times$ & 597.52 & 6.22$\times$
& \textbf{28.71} & \textbf{0.6002} & \textbf{0.4692} & \textbf{0.9451} \\

\textbf{SpeCa} $(100,1,6,20)$
& 40.02 & 2.84 & 3.27$\times$ & 605.23 & 6.15$\times$
& 28.43 & 0.6111 & 0.4867 & 0.8874 \\
\rowcolor{gray!20}
\textbf{SpeCa + Z-Cache} $(100,1,6,20)$
& 40.02 & 3.04 & 3.05$\times$ & 606.27 & 6.14$\times$
& \textbf{28.67} & \textbf{0.6327} & \textbf{0.4420} & \textbf{0.9004} \\

\textbf{MagCache} $(.80,7,.02)$
& 33.85 & 1.86 & 5.00$\times$ & 596.98 & 6.23$\times$
& 28.88 & 0.6582 & 0.4275 & 0.9415 \\
\rowcolor{gray!20}
\textbf{MagCache + Z-Cache} $(.80,7,.02)$
& 37.98 & 2.78 & 3.34$\times$ & 598.12 & 6.22$\times$
& \textbf{29.05} & \textbf{0.6710} & \textbf{0.3994} & \textbf{0.9511} \\

\bottomrule
\end{tabular}}
\label{tab:full_paired_base_predictors}
\end{table*}

%% file: tab/HunyuanVideo-Metrics.tex
\begin{table*}[h]
\centering
\caption{\textbf{Quantitative comparison on text-to-video generation} with HunyuanVideo on VBench. L-Speed and F-Speed are the latency and FLOPs speedups relative to the original 50-step model. Within each caching block, bold numbers indicate the best values, including ties.}

\setlength\tabcolsep{5.0pt} 
\small
\resizebox{\linewidth}{!}{
\begin{tabular}{l | c | c  c | c  c | c }
    \toprule
    {\bf Method} & {\bf Efficient} &\multicolumn{4}{c|}{\bf Acceleration} &{\bf VBench $\uparrow$}  \\
    \cline{3-6}
    {\bf (on HunyuanVideo)} & {\bf Attention } & {\bf Latency(s) $\downarrow$} & {\bf L-Speed $\uparrow$} & {\bf FLOPs(T) $\downarrow$}  & {\bf F-Speed $\uparrow$} & \bf Score(\%) \\
    \midrule

  $\textbf{Original: 50 steps}$ 
                           & \ding{52}  &  145.00 & 1.00$\times$& {29773.0}   & {1.00$\times$} & {80.66}      \\ 
  {$22\%$\textbf{ steps}}  & \ding{52}  &  31.87 & 4.55$\times$ & {6550.1}   & {4.55$\times$} & {78.74}          \\
                           
  \midrule

\textbf{TeaCache}$({l}=0.4)$ & \ding{52} & 26.61 & 5.45$\times$ & 6550.1 & 4.55$\times$ & 79.36  \\

$\textbf{FORA}(N=5)$ 
& \ding{52}  &  34.39 & 4.22$\times$  & \textbf{5960.4}   & \textbf{5.00$\times$} & {78.83}     \\

$\textbf{\texttt{ToCa}}$ $(\mathcal{N}=5,R=90\%)$

                           & \ding{56}  &  38.52 & 3.76$\times$ & {7006.2}   & {4.25$\times$} & {78.86}    \\
  
$\textbf{\texttt{DuCa}} $ $(\mathcal{N}=5,R=90\%)$ 

                           & \ding{52}  &  31.69 & 4.58$\times$ & {6483.2}   & {4.59$\times$} & {78.72}    \\

$\textbf{TaylorSeer} $ $(\mathcal{N}=5,O=1)$ 
                           & \ding{52}  & \textbf{25.34} & \textbf{5.72$\times$} & \textbf{5960.4}  & \textbf{5.00$\times$} &79.93 \\

\rowcolor{gray!20}
$\textbf{Z-Cache} $ $(\mathcal{N}=5,O=1)$ 
                           & \ding{52}  & 28.69 & $5.05\times$ & 5997.2  & $4.96\times$ &\textbf{80.10}  \\

    \midrule

\textbf{TeaCache}$({l}=0.5)$ & \ding{52} & 26.61 & 5.45$\times$ & 5359.1 & 5.56$\times$ & 78.32   \\

$\textbf{TaylorSeer} $ $(\mathcal{N}=7,O=1)$ 
                           & \ding{52}  & \textbf{21.01} & \textbf{6.90}$\times$ & \textbf{4794.5}  & \textbf{6.21}$\times$ &{79.09}   \\

\rowcolor{gray!20}
$\textbf{Z-Cache} $ $(\mathcal{N}=7,O=1)$ 
                           & \ding{52}  & 23.08 & $6.28 \times$ & 4799.6  & $6.20\times$ &\textbf{79.60}   \\

    \bottomrule
\end{tabular}}
\label{table:HunyuanVideo-Metrics}
\vspace{-0.2cm}
\end{table*}

%% file: sec/conclusion.tex
\section{Conclusion}
This paper introduces \textit{Z-Cache}, a novel caching framework that redefines the efficiency-quality trade-off for diffusion transformers. Central to its innovation is a self-reflective mechanism, which periodically refines intermediate features through "future-aware" feature integration—enabling substantial acceleration without sacrificing output fidelity. 
By integrating this mechanism, \textit{Z-Cache} achieves exceptional acceleration (up to 6.22× faster inference on FLUX-dev.1) while preserving high output quality. Beyond static image generation, Z-Cache demonstrates remarkable adaptability to video tasks, attaining a state-of-the-art 6.20× acceleration rate for videos, all while maintaining high output quality.

%% file: sec/appendix.tex
\section{Appendix}
\subsection{Additional Related Works}
\label{app:related_full}

\paragraph{Diffusion models and diffusion transformers.}
Diffusion models~\citep{sohl2015deep,ho2020DDPM} generate data through iterative denoising and have become a dominant framework for visual generation. Early systems were largely based on U-Net architectures~\citep{ronneberger2015unet}, while recent large-scale models increasingly adopt diffusion transformers (DiTs)~\citep{peebles2023dit}. This architectural shift has enabled strong image and video generation performance across modern systems~\citep{chen2023pixartalpha,chen2024pixartsigma,opensora,yang2025cogvideox}. However, the iterative sampling process remains computationally expensive, motivating two broad acceleration directions: reducing the number of sampling timesteps and reducing the per-step cost of the denoising network.

\paragraph{Sampling timestep reduction.}
Timestep reduction methods aim to preserve generation quality with fewer denoising iterations. DDIM~\citep{songDDIM} introduces deterministic sampling for faster inference, while DPM-Solver and its variants~\citep{lu2022dpm,lu2022dpm++,zheng2023dpmsolvervF} use high-order ODE solvers to reduce the number of function evaluations. Rectified Flow~\citep{refitiedflow} learns straighter paths between noise and data, and distillation-based approaches compress multi-step generation into fewer steps~\citep{salimans2022progressive,meng2022on}. Consistency Models~\citep{song2023consistency} further enable one-step or few-step generation by directly mapping noisy inputs to clean outputs. These methods reduce sampling length and are orthogonal to our goal of improving feature caching within a given sampling schedule.

\paragraph{Denoising network acceleration.}
Another line of work accelerates the denoising network at each timestep. Compression-based methods include pruning~\citep{structural_pruning_diffusion,zhu2024dipgo}, quantization~\citep{10377259,shang2023post,kim2025ditto}, distillation~\citep{li2024snapfusion}, and token reduction or token merging~\citep{bolya2023tomesd,kim2024tofu,zhang2024tokenpruningcachingbetter,zhang2025sito,cheng2025catpruningclusterawaretoken,saghatchian2025cached}, alongside dynamic-resolution approaches~\citep{zheng2026sketchfrescoefficientdiffusion,yan2026specedittrainingfreeaccelerationdiffusion,zheng2026dynamicvideogenerationshaping,qin2026avitsadaptivespatiotemporaltoken}. These approaches reduce computation by simplifying model weights, activations, or token sequences. Although effective, they often require tuning, calibration, or careful engineering to maintain the quality--efficiency balance. Z-Cache is complementary to these methods because it is training-free and modifies the inference-time feature-caching trajectory rather than the model parameters.

\paragraph{Feature caching-based acceleration.}
Feature caching exploits temporal redundancy across diffusion timesteps to accelerate inference~\citep{surveycachemethodsdiffusion,liu2025freqcaacceleratingdiffusionmodels}. Early caching methods for U-Net diffusion models include Faster Diffusion~\citep{li2023FasterDiffusion} and DeepCache~\citep{ma2024deepcache}. Recent methods extend this idea to transformer-based diffusion models. FORA~\citep{selvaraju2024fora} and $\Delta$-DiT~\citep{chen2024delta-dit} reuse attention or representation states across nearby steps. TeaCache~\citep{liu2024timestep} uses timestep-adaptive caching, while ToCa and DuCa~\citep{zou2024accelerating,zou2024DuCa} cache token-wise or dual feature representations. Other methods improve efficiency through error-aware caching, adaptive caching-pruning combinations, or spatially varying sampling strategies~\citep{qiu2025acceleratingdiffusiontransformererroroptimized,sun2025unicpunifiedcachingpruning,liu2025regionadaptivesamplingdiffusiontransformers}.

\paragraph{Forecasting and predictor-oriented caching.}
More recent feature-caching methods improve the base cache predictor itself~\citep{zheng2025letfeaturesdecidesolvers,cai2026lesalearnablestageawarepredictors,shen2026acceleratingdiffusiontransformersgaussian,zou2026discaacceleratingvideodiffusion,shen2026fixedformulasdatadrivenlinear}. TaylorSeer~\citep{liu2025reusingforecastingacceleratingdiffusion} predicts intermediate features using polynomial extrapolation from neighboring cached activations, which establishes the \textit{cache-then-forecast} paradigm. SpeCa~\citep{Liu_2025} introduces speculative feature caching, and MagCache~\citep{ma2025magcachefastvideogeneration} exploits magnitude-aware caching for efficient generation. These works primarily focus on improving feature prediction or cache scheduling.

\paragraph{Sensitivity-aware, second-order, and spatial residual caching.}
Recent methods address different sources of caching error. SenCache~\citep{haghighi2026sencache} estimates how denoising outputs respond to changes in the latent input and timestep, and uses these sensitivities to decide whether cached outputs can be reused for each sample. Its sensitivity estimates are obtained using a small calibration set. D2Cache~\citep{liu2026d2cache} augments a cached first-order residual with a second-order residual difference and scales this correction using error estimates derived from timestep embeddings. ResCa~\citep{fang2026resca} exploits similarity among token denoising trajectories: it computes a proxy token for each cluster and uses multi-order residuals to guide simulated updates of the remaining tokens. Its updates combine a token's historical residuals with current residual information from the proxy. These approaches respectively emphasize sensitivity-guided cache decisions, temporal residual prediction, and spatially shared denoising information.

\paragraph{Relation to Z-Cache.}
Z-Cache performs retrospective trajectory correction within an active cache window. Once the next scheduled full-computation anchor is observed on the provisional trajectory, Z-Cache combines past-anchor and future-anchor predictions, replays the window, and commits the reflected endpoint before continuing denoising. This differs from SenCache's reuse decisions, D2Cache's temporal extrapolation, and ResCa's current-step proxy-guided token updates. Z-Cache can therefore complement compatible cache policies and predictors while preserving the base full-computation schedule, provided their cached representations and state updates support dual-anchor prediction and window replay.

\subsection{Experiment Settings}

\subsubsection{DiT} 

We conducted experiments mainly on three state-of-the-art diffusion architectures: DiT-XL/2 \citep{peebles2023dit}, FLUX.1-dev \citep{flux2024}, and HunyuanVideo \citep{kong2024hunyuanvideo}. A per-model subset of baselines was systematically selected from the literature: FORA \citep{selvaraju2024fora}, ToCa \citep{zou2024accelerating}, DuCa \citep{zou2024DuCa}, TeaCache \cite{liu2024timestep}, and TaylorSeer \citep{liu2025reusingforecastingacceleratingdiffusion}.

\subsubsection{FLUX-dev.1} 
We evaluate text-guided image generation on FLUX.1-dev with DrawBench. Each
method generates 200 images at $1024\times1024$ resolution using the default
sampling configuration. To make the efficiency comparison explicit, we report
end-to-end H100 latency, peak CUDA memory, and FLOPs under the same
implementation. FLOPs speed is normalized by the original 50-step FLUX.1-dev
cost. PSNR, SSIM, and LPIPS are computed against the corresponding full
50-step FLUX.1-dev outputs under the same prompts and seeds, while
ImageReward is computed on generated image--prompt pairs.

\subsubsection{HunyuanVideo} 

For video generation tasks, all experiments utilized the model's default configuration. We generated a total of 4720 video sequences (65 frames per sequence at 640$\times$480 resolution) from VBench with 4 samples each. Evaluation employed the comprehensive VBench framework. Since all methods produce noticeable semantic shifts, we skipped all pixel-level evaluation for the best interpretation.

\input{sec/zcache_rebuttal_experiments}

\input{sec/zcache_pseudocode}

\subsection{Impact Statement}
This work improves the efficiency of diffusion-model inference via a training-free caching-and-reflection strategy (\textit{Z-Cache}). By reducing the compute and latency required per generated sample, it can lower energy use and broaden access to high-quality generative models for research and practical applications. However, cheaper and faster generation may also enable misuse (e.g., large-scale misinformation, deceptive or non-consensual synthetic media, and spam) and can amplify existing concerns such as bias or copyright infringement. \textit{Z-Cache} does not change the underlying model’s training data or safety properties, so responsible deployment should rely on standard safeguards such as access controls, rate limiting, moderation/filtering, and provenance mechanisms where appropriate.

%% file: sec/zcache_rebuttal_experiments.tex
\subsection{Supplementary Experiments}
\label{app:supplementary_experiments}
\subsubsection{Global trajectory comparison}
\label{app:global_trajectory}
We compare Full sampling, TaylorSeer without reflection, and canonical
Z-Cache on FLUX.1-dev using 32 frozen DrawBench prompts, two seeds per prompt,
and cache intervals $N\in\{6,8\}$. Each paired sample shares its initial
latent and the 50-step scheduler. We save all 50 latent tensors. For Z-Cache,
the last replayed value replaces the provisional value at each revisited
step; this reconciled trajectory is the primary comparison.

For method $M$, define the relative latent error and its normalized
trapezoidal area under the curve (AUC) as
\begin{equation}
 d_i^M=\frac{\lVert x_{i+1}^M-x_{i+1}^{\mathrm{Full}}\rVert_2}
 {\max(\lVert x_{i+1}^{\mathrm{Full}}\rVert_2,10^{-12})},
 \qquad
 A^M=\frac{1}{T-1}\sum_{i=0}^{T-2}\frac{d_i^M+d_{i+1}^M}{2},\quad T=50.
\end{equation}
Table~\ref{tab:global_trajectory_rebuttal} reports sample means.

\begin{table}[H]
\centering
\small
\setlength{\tabcolsep}{3pt}
\caption{Global latent trajectories relative to the shared Full reference.}
\label{tab:global_trajectory_rebuttal}
\begin{tabular}{clrr}
\toprule
$N$ & Metric & TaylorSeer & Z-Cache \\
\midrule
6 & AUC $\downarrow$ & 0.212977 & \textbf{0.198117} \\
6 & Final rel.-$L_2$ $\downarrow$ & 0.574011 & \textbf{0.546836} \\
6 & Final cosine $\uparrow$ & 0.823749 & \textbf{0.841404} \\
\midrule
8 & AUC $\downarrow$ & 0.256287 & \textbf{0.235652} \\
8 & Final rel.-$L_2$ $\downarrow$ & 0.669091 & \textbf{0.630887} \\
8 & Final cosine $\uparrow$ & 0.762938 & \textbf{0.792403} \\
\bottomrule
\end{tabular}
\end{table}

Reflection reduces mean trajectory AUC by 6.98\% and 8.05\% at $N=6$ and
$N=8$, respectively. All three prompt-averaged metrics improve for all
32 prompts at both settings. This does not imply uniform sample-wise
improvement: final relative $L_2$ improves for 60/64 and 63/64 samples,
respectively. All 52,736 checked trajectory values are finite and no rank
fails. These are latent-trajectory measurements, not additional image-quality
or latency scores, and they do not establish a universal stability or
convergence guarantee.

\subsubsection{Same-start local reflection analysis}
\label{app:local_trajectory}
To separate the effect of reflection from previously accumulated drift,
we fork three branches from the exact same committed canonical Z-Cache
left-anchor state $x_{a+1}$: a local Full branch, the provisional TaylorSeer branch,
and canonical reflected replay. The local Full branch evaluates the full
network at every post-anchor step. It supplies neither features nor the
future anchor to the cached branches and is a diagnostic reference, not an
equal-compute baseline. The experiment uses the same 32 prompts and two
seeds as Section~\ref{app:global_trajectory}, with seven windows per sample
at $N=6$ and five at $N=8$, giving 768 matched windows.

We first average window metrics within each sample. For the endpoint-displacement
intervals reported below, we use 95\% percentile bootstrap intervals from
20,000 prompt-cluster resamples, retaining both seeds of each sampled prompt.
Local AUC is the normalized trapezoidal relative-$L_2$ error from the common
start through the window endpoint $x_{b+1}$. Table~\ref{tab:local_trajectory_rebuttal} uses the conditional
local Full trajectory as its reference; it must not be pooled with the
global Full comparison.

\begin{table}[H]
\centering
\small
\setlength{\tabcolsep}{3pt}
\caption{Same-start local trajectories relative to the conditional local
Full reference.}
\label{tab:local_trajectory_rebuttal}
\begin{tabular}{clrr}
\toprule
$N$ & Metric & TaylorSeer & Z-Cache \\
\midrule
6 & AUC $\downarrow$ & 0.012688 & \textbf{0.008928} \\
6 & Endpoint rel.-$L_2$ $\downarrow$ & 0.026057 & \textbf{0.016922} \\
6 & Endpoint cosine $\uparrow$ & 0.999620 & \textbf{0.999828} \\
\midrule
8 & AUC $\downarrow$ & 0.020822 & \textbf{0.015519} \\
8 & Endpoint rel.-$L_2$ $\downarrow$ & 0.045693 & \textbf{0.031364} \\
8 & Endpoint cosine $\uparrow$ & 0.998882 & \textbf{0.999430} \\
\bottomrule
\end{tabular}
\end{table}

All three sample-aggregated local metrics favor reflection for 64/64 samples
and 32/32 prompts at both intervals. The common starts match exactly and all
20,992 checked local relative-$L_2$/cosine values are finite.

The replayed endpoint intentionally differs from the provisional endpoint.
Its direct relative-$L_2$ displacement, normalized by the provisional
TaylorSeer endpoint norm, is 0.014184 [0.013653, 0.014733] at $N=6$ and
0.023060 [0.022209, 0.023909] at $N=8$ (sample means and prompt-cluster
95\% intervals). The corresponding window-level 95th percentiles are
0.025632 and 0.037227. Thus, endpoint changes are measurable; the same-start
control shows that they move the tested local trajectories closer to
conditional Full, without asserting that every possible trajectory or
cache predictor behaves similarly.

\subsubsection{Fusion rules and replay implementation}
\label{app:fusion_backend}
We evaluate four reflection functions on 200 paired DrawBench prompts with
FLUX.1-dev, TaylorSeer $N=6$, and the same frozen Base schedule. With Taylor
order $K$, distances $d_L=i-a$, $d_R=b-i$, and $r=d_L/(d_L+d_R)$, the weights
on the future-conditioned feature prediction are
\begin{equation}
\begin{split}
w_{\rm linear}&=r,\qquad
w_{\rm remainder}=\frac{d_L^{K+1}}{d_L^{K+1}+d_R^{K+1}},\\
w_{\rm variance}&=\frac{d_L^{2(K+1)}}{d_L^{2(K+1)}+d_R^{2(K+1)}},\qquad
w_{\rm smoothstep}=r^2(3-2r).
\end{split}
\end{equation}
The ``variance-optimal'' name refers to an independent-error model using
squared Taylor-remainder scales; it does not use oracle errors.

\paragraph{Image-evaluation protocol.}
Tables~\ref{tab:fusion_rules_rebuttal}--\ref{tab:recent_cache_rebuttal}
report SSIM computed on grayscale images; LPIPS and ImageReward use the
same frozen evaluation configuration.

\begin{table}[H]
\centering\small
\caption{Fusion-function ablation. Bold marks the best point estimate in
each SSIM, LPIPS, and IR column; no confidence intervals or human-preference study establish
quality significance.}
\label{tab:fusion_rules_rebuttal}
\begin{tabular}{lrrr}
\toprule
Fusion rule & SSIM$\uparrow$ & LPIPS$\downarrow$ & IR$\uparrow$ \\
\midrule
Original linear & 0.6462 & 0.3914 & \textbf{0.9908} \\
Remainder ratio & 0.6474 & 0.3888 & 0.9763 \\
Variance-optimal & \textbf{0.6476} & \textbf{0.3883} & 0.9863 \\
Jet smoothstep & 0.6471 & 0.3898 & 0.9821 \\
\bottomrule
\end{tabular}
\end{table}

All four rules use the same Base schedule, 747.501934 TFLOPs, and approximately
41.652 GiB peak allocated memory. Variance-optimal leads SSIM and LPIPS,
whereas linear leads ImageReward. Linear remains a simple parameter-free
default.

\begin{table}[H]
\centering\small
\setlength{\tabcolsep}{3pt}
\caption{Two replay backends under the same TaylorSeer $N=6$ schedule.
Image quality uses DrawBench200; efficiency is a separate frozen-prompt
A100 diagnostic.}
\label{tab:backend_rebuttal}
\begin{tabular}{lrrr}
\toprule
Backend & SSIM$\uparrow$ & LPIPS$\downarrow$ & IR$\uparrow$ \\
\midrule
Block-wise linear & 0.6462 & 0.3914 & 0.9908 \\
Root-residual state graft & \textbf{0.6505} & \textbf{0.3797} & \textbf{0.9916} \\
\midrule
Backend & Latency (s)$\downarrow$ & TFLOPs$\downarrow$ & Peak GiB$\downarrow$ \\
\midrule
Block-wise linear & 6.7656 & 747.502 & 41.652 \\
Root-residual state graft & \textbf{5.4130} & \textbf{747.095} & \textbf{40.011} \\
\bottomrule
\end{tabular}
\end{table}

The graft reflects root residuals and bypasses the 57 Transformer blocks
during replay instead of fusing 95 internal feature sites. Both backends
retain 10 primary FULL steps, 40 primary PREDICT steps, seven closed windows,
and zero extra endpoint FULL calls. Latency is the median of seven balanced
rounds after one excluded warm-up on one A100 and frozen prompt 0, including
hooks, denoising, and replay but excluding model loading, prompt encoding,
latent preparation, and VAE decoding. Memory is maximum CUDA allocated memory;
FLOPs are a separate exact-trajectory Transformer-forward profile.
The graft reduces latency by 19.99\% and peak memory by 3.94\%, while changing
the approximation rather than preserving identical outputs. This is not an
H100 reproduction or evidence that overhead is eliminated for other Bases,
intervals, or models.

\subsubsection{Composition with recent cache predictors}
\label{app:recent_cache_experiments}
We evaluate audited clean-room FLUX ports of SenCache and D2Cache on all
200 DrawBench prompts. Within each Base/+Z pair, prompts, seeds, initial
latents, and the Base FULL schedule are identical. These controlled ports
are additional composability evidence, not measurements of the authors'
official implementations. Table~\ref{tab:recent_cache_rebuttal} uses the
image-evaluation protocol described in
Section~\ref{app:fusion_backend}; bold indicates the better SSIM, LPIPS, or ImageReward point estimate within
each matched pair.

\begin{table}[H]
\centering\small
\setlength{\tabcolsep}{2.5pt}
\caption{Matched recent-predictor results. FULL is the mean number of primary
FULL evaluations across 200 samples. F-Speed is measured on one frozen
prompt-0 trajectory; it is not a mean over 200 prompts or a latency speedup.}
\label{tab:recent_cache_rebuttal}
\begin{tabular}{lclrrrrr}
\toprule
Base & $N$ & Arm & FULL & F-Speed & SSIM & LPIPS & IR \\
 & & & & $\uparrow$ & $\uparrow$ & $\downarrow$ & $\uparrow$ \\
\midrule
SenCache & 6 & Base & 10.000 & 4.996 & 0.6115 & 0.5126 & 0.7210 \\
 & 6 & +Z & 10.000 & 4.987 & \textbf{0.6149} & \textbf{0.4917} & \textbf{0.7985} \\
\midrule
SenCache & 7/8 & Base & 8.000 & 6.243 & 0.5943 & 0.5559 & 0.5486 \\
 & 7/8 & +Z & 8.000 & 6.230 & \textbf{0.5982} & \textbf{0.5289} & \textbf{0.6256} \\
\midrule
D2Cache & 6 & Base & 10.045 & 4.996 & 0.6488 & 0.4211 & \textbf{0.9536} \\
 & 6 & +Z & 10.045 & 4.987 & \textbf{0.6515} & \textbf{0.4158} & 0.9415 \\
\midrule
D2Cache & 7 & Base & 9.015 & 5.550 & 0.6346 & 0.4416 & 0.9411 \\
 & 7 & +Z & 9.015 & 5.539 & \textbf{0.6358} & \textbf{0.4390} & \textbf{0.9554} \\
\midrule
D2Cache & 8 & Base & 8.140 & 6.243 & \textbf{0.6146} & \textbf{0.4775} & 0.8706 \\
 & 8 & +Z & 8.140 & 6.229 & 0.6134 & 0.4778 & \textbf{0.9107} \\
\bottomrule
\end{tabular}
\end{table}

SenCache improves SSIM, LPIPS, and ImageReward at each distinct operating point. Its $N=7$ calibration
target realizes eight FULL evaluations, yielding the same schedules and
pixel-identical outputs as $N=8$ for each arm; we combine these rows rather
than count them as independent replications. D2Cache is mixed: at $N=6$,
SSIM/LPIPS improve while ImageReward decreases; at $N=8$, only ImageReward
improves among these three metrics. No quality confidence intervals or human-preference
evaluation support a significance or uniform-dominance claim.

F-Speed is the 50-step Full Transformer FLOPs divided by each arm's
whole-trajectory Transformer FLOPs, including reflection forwards. The
measured additional compute is 0.17--0.21\% for these five distinct operating
points. The separate A100 latency measurements are preliminary execution
diagnostics and are omitted because they do not match the original H100
hardware and optimized timing protocol. Low additional Transformer FLOPs
therefore do not imply universally low wall-clock or memory overhead.

%% file: sec/zcache_pseudocode.tex
\subsection{Pseudocode and State Updates}
\label{app:zcache_pseudocode}

Algorithm~\ref{alg:zcache_window} specifies the TaylorSeer-based block-feature
instantiation with linear fusion used in the trajectory experiments, with a
deterministic step-indexed scheduler. Let $a<b$ be consecutive indices selected for
FULL computation by the base schedule. The saved left boundary is $x_{a+1}$.
The notation $x_i$, $F_i$, and $\mathcal H_i$ follows the main text.

Let $\mathcal B^p$ be the frozen predictor state at the left boundary and
$\mathcal C(\mathcal B,i)$ its feature prediction.
$F_i(x,\mathcal H)$ evaluates the uncached network components and the scheduler
using the supplied cache-site features.
$\operatorname{FullStep}_i(x)$ performs one full network evaluation and its
scheduler update, returning the output latent and the observed feature bank.
Both operators use the original timestep and step index $i$; replay does not
advance a shared scheduler cursor.

\begin{algorithm}[H]
\caption{Z-Cache: forecast, reflect, and commit one window}
\label{alg:zcache_window}
\small
\begin{algorithmic}[1]
\REQUIRE Scheduled FULL indices $a<b$; saved latent $x_{a+1}$;
         frozen history $\mathcal B^p$; predictor $\mathcal C$
\ENSURE Committed latent $\tilde x_{b+1}$ and predictor state $\mathcal B_b$
\STATE $\bar x_{a+1}\gets x_{a+1}$ \COMMENT{provisional trajectory}
\FOR{$i=a+1,\ldots,b-1$}
  \STATE $\hat{\mathcal H}_i^p\gets\mathcal C(\mathcal B^p,i)$
  \STATE $\bar x_{i+1}\gets F_i(\bar x_i,\hat{\mathcal H}_i^p)$
\ENDFOR
\STATE $(\bar x_{b+1},\mathcal H_b)\gets\operatorname{FullStep}_b(\bar x_b)$
       \COMMENT{the scheduled right FULL}
\STATE $\mathcal B^f\gets\operatorname{Update}(\mathcal B^p,b,\mathcal H_b)$
\IF{$b=a+1$}
  \STATE \textbf{return} $(\bar x_{b+1},\mathcal B^f)$ \COMMENT{no cached interior}
\ENDIF
\STATE $\tilde x_{a+1}\gets x_{a+1}$ \COMMENT{restore the left boundary}
\FOR{$i=a+1,\ldots,b$}
  \STATE $w_i\gets(i-a)/(b-a)$
  \STATE $\tilde{\mathcal H}_i\gets
    (1-w_i)\mathcal C(\mathcal B^p,i)+w_i\mathcal C(\mathcal B^f,i)$
  \STATE $\tilde x_{i+1}\gets F_i(\tilde x_i,\tilde{\mathcal H}_i)$
         \COMMENT{cached replay, including step $b$}
\ENDFOR
\STATE $\mathcal B_b\gets\operatorname{Update}(\mathcal B^p,b,\tilde{\mathcal H}_b)$
       \COMMENT{reconcile the closing cache}
\STATE \textbf{return} $(\tilde x_{b+1},\mathcal B_b)$
\end{algorithmic}
\end{algorithm}

\paragraph{Cache reconciliation.}
$\operatorname{Update}(\mathcal B^p,b,\mathcal H)$ rebuilds the predictor state
from the frozen past state and the supplied endpoint features; for TaylorSeer,
it rebuilds the finite-difference coefficients at $b$. At each cache site, let
$D_a^0,\ldots,D_a^{K_a}$ be the available coefficients and $q$ the maximum
order. The update is
\[
  D_b^0=\mathcal H,\qquad
  D_b^k=\frac{D_b^{k-1}-D_a^{k-1}}{b-a},
  \quad k=1,\ldots,\min(q,K_a+1).
\]
Both predictor states remain fixed during replay.
With the linear weight and Taylor prediction, $w_b=1$ and
$\tilde{\mathcal H}_b=\mathcal H_b$.
This bank was observed on the provisional input to step $b$; it is not a new
FULL evaluation on the replayed input. The replayed output latent can therefore
differ from $\bar x_{b+1}$.

\paragraph{Sampling-loop integration.}
Execute the initial FULL step before opening a window, then use each returned
$(\tilde x_{b+1},\mathcal B_b)$ as the next left boundary. Warm-up FULL steps follow the
base schedule. Every scheduled closing step is completed even if no later
closed window remains. A cached tail with no scheduled right FULL follows the base
predictor without reflection; no extra endpoint FULL is inserted.
The procedure preserves the FULL schedule while adding cached replay and
scheduler operations.